\documentclass{article}
\usepackage{ijcai19}

\usepackage{times}
\usepackage{soul}
\usepackage{url}
\usepackage[hidelinks]{hyperref}
\usepackage[utf8]{inputenc}
\usepackage[small]{caption}
\usepackage{graphicx}
\usepackage{amsmath}
\usepackage{booktabs}
\usepackage{algorithm}
\usepackage{algorithmic}

\usepackage{amsfonts}
\usepackage{amsmath}

\title{GCN-LASE: Towards Adequately Incorporating Link Attributes in Graph Convolutional Networks \thanks{This work was supported by the National Natural Science Foundation of China (Grant No. 61876006 and No. 61572041).}}

\author{
Ziyao Li$^1$ \and
Liang Zhang$^1$ \And
Guojie Song$^2$\thanks{Corresponding author.}
\affiliations
$^1$Yuanpei College, Peking University, China\\
$^2$Key Laboratory of Machine Perception, Ministry of Education, Peking University, China
\emails
\{leeeezy, zl505, gjsong\}@pku.edu.cn
}

\begin{document}

\maketitle
\frenchspacing

\begin{abstract}
  Graph Convolutional Networks (GCNs) have proved to be a most powerful architecture in aggregating local neighborhood information for individual graph nodes. Low-rank proximities and node features are successfully leveraged in existing GCNs, however, attributes that graph links may carry are commonly ignored, as almost all of these models simplify graph links into binary or scalar values describing node connectedness. In our paper instead, links are reverted to hypostatic relationships between entities with descriptional attributes. We propose GCN-LASE (GCN with Link Attributes and Sampling Estimation), a novel GCN model taking both node and link attributes as inputs. To adequately captures the interactions between link and node attributes, their tensor product is used as neighbor features, based on which we define several graph kernels and further develop according architectures for LASE. Besides, to accelerate the training process, the sum of features in entire neighborhoods are estimated through Monte Carlo method, with novel  sampling strategies designed for LASE to minimize the estimation variance. Our experiments show that LASE outperforms strong baselines over various graph datasets, and further experiments corroborate the informativeness of link attributes and our model's ability of adequately leveraging them.
\end{abstract}

\section{Introduction}

\begin{figure*}
  \centering
  \includegraphics[width=0.91\textwidth]{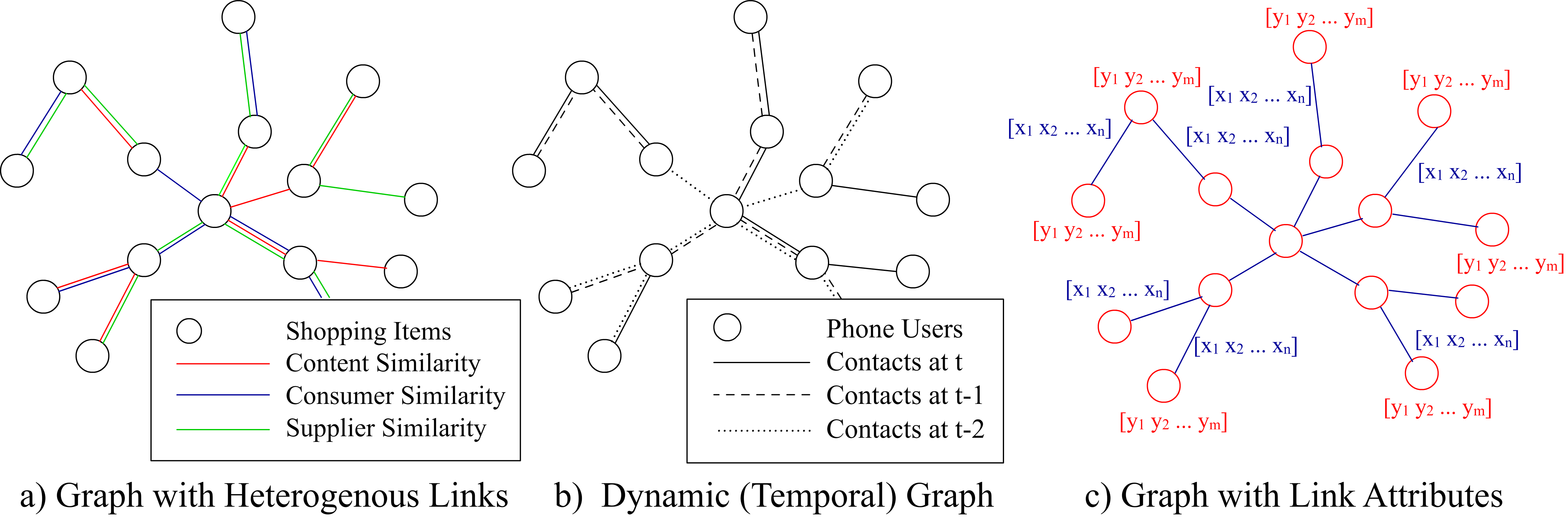}
  \caption{Different types of graphs that LASE can deal with: {\bf a)} commodity network: an example of graphs with heterogenous links; {\bf b)} contact network: an example of temporal graph; {\bf c)} graph with both node and link attributes. By arranging link weights from different perspectives or time-stamps into vectors, it is easy to turn {\bf a)} {\bf b)} into {\bf c)}.}\label{fig1}
\end{figure*}

  After several attempts and modifications (\cite{bruna-gcn} \cite{ger-gcn} \cite{kipf-gcn} \cite{ker2nn} \cite{fastgcn}), Graph Convolutional Networks (GCNs) \footnote{Although some papers do not explicitly use the term {\it convolution}, such as \cite{ker2nn} and \cite{sage}, a similar idea of aggregating features in a node's neighborhood can be seen in the models. Therefore, we categorize all similar work into the class of graph convolutional networks.} are rapidly gaining popularity due to its excellent performance in aggregating neighborhood information for individual graph nodes. Although low-rank proximities and node features are successfully leveraged through the convolutional layers, the attributes that graph links may carry are generally ignored in GCNs. In almost all existing GCNs, graph links are regarded as indicators of proximities between nodes. These proximities are only used to identify neighborships (and their influences if weighted) in the local neighborhoods.

  However, in real-world scenarios, a link between a pair of nodes carries a lot more information than a simple indicator of neighborship. It represents a {\it hypostatic relationship} between two entities, usually with concrete attributes. For example, two connected people in a social network may have different relationships including family members, colleagues and alumni, and thus their communication patterns and contents may vary; a link in a business network represents a transaction between two companies, and the properties of such transactions are tremendously informative. Therefore, revisiting these link attributes, which are generally ignored in current GCNs, allows us to recover the exact relationships between nodes.

  While one can leverage link attributes in GCNs with tricks, such as concatenating them to neighbor node attributes, these implementations cannot adequately captures the interactions between the attributes \footnote{See Section 3 for detailed introduction.}. To the best of our knowledge, there is no previous work focusing on incorporating link attributes into GCNs. We propose {\bf GCN-LASE (Graph Convolutional Network with Link Attributes and Sampling Estimation)} as an attempt. LASE is an extension of GCN, which also learns a function that maps a target node to its hidden representation considering the local neighborhood features and structures, but including both link and neighbor node attributes. The aggregated features are then used to conduct downstream tasks.

  To leverage the link and node attributes as well as the {\it interactions} in between, we adopt their {\it tensor products} as the fully associated neighbor features, based on which a {\it neighbor kernel} is developed using the {\it inner products of tensors}. We further derive corresponding graph kernels and finally the neural architectures following a similar route introduced in \cite{ker2nn}. An intuitive understanding of LASE is then demonstrated, by modularizing it into a {\it gate}, an {\it amplifier}, and an {\it aggregator}. Meanwhile, we adopt the Monte Carlo method to quickly estimate the sum of features in neighborhoods. We also introduce a novel sampling setup for LASE to reduce the estimation variance.

  Recovering more information is not the only benefit LASE brings about--it also enlarges the expressive abilities of graph structures that GCNs can handle. At lease two other types of graph-structured data can be handled with LASE: i) Graphs with heterogenous links, where link weights from different perspectives can be arranged into vectors and used as attributes; ii) Dynamic or temporal graphs, where link weights from different time-stamps can be stacked together and used as attributes. See Figure~\ref{fig1} for detailed examples. We validate our approach on four datasets across different domains, and further design additional experiments in order to demonstrate the informativeness of link attributes and the effect of our sampling strategy.

\section{Related Work}

  Our method builds upon previous work of machine learning on graph data, including graph convolutional networks, graph kernels and node representation learning.

  \paragraph{Graph convolutional networks.} The past few years have seen plenty of works focusing on implementing deep architectures over graphs (\cite{bruna-gcn} \cite{henaff-gcn} \cite{kipf-gcn} \cite{sdne} \cite{ger-gcn} \cite{sage} \cite{ker2nn}), among which the convolutional networks seem to be the most appealing. There are mainly two types of convolutional networks existing, one learning features for entire graphs (\cite{ker2nn} \cite{ger-gcn}), the other for individual nodes (\cite{kipf-gcn} \cite{sage}). Both methods interpret the concept of {\it convolution} as merging node features in local neighborhoods. Meanwhile, as early GCNs (\cite{kipf-gcn}) does not support mini-batch training, modifications towards better efficiency emerge (\cite{sage} \cite{fastgcn} \cite{ada-sample}), in which the Monte Carlo method is generally used to estimate neighborhood representations through a controllable size of nodes. In our paper, a different sampling implementation is adopted, which makes a tradeoff between variance and efficiency by controlling the interval of calculating the optimal sampling probabilities.

  \paragraph{Graph kernels.} Kernel methods \cite{kernel} have long been an important class of machine learning techniques, while it remains challenging to define effective and convenient kernels for graphs. Existing graph kernels (\cite{rw-kernel} \cite{g-kernels} \cite{wl-kernel}) are typically defined over of sub-structures in graphs, such as sub-trees and walk sequences. Later, \cite{ker2nn} introduces an innovative route to develop neural architectures using graph kernels\footnote{These architectures have better explainability as the hidden representations can be interpreted as a series of inner products of the input graph and some sample graphs. See Theorem 1 for detailed explanation.}.  To the best of our knowledge, there is no existing work aiming at incorporating link attributes into graph kernels. In our paper, we adopt a similar route as \cite{ker2nn} to design LASE, but using novel graph kernels which are able to handle link attributes.

  \paragraph{Node representation learning (NRL).} NRL aims to learn low-dimensional embeddings for graph nodes (\cite{dw} \cite{line} \cite{n2v} \cite{struct2vec} \cite{sage} \cite{sepne} \cite{dne}), which are later used in downstream prediction tasks such as node classification and link prediction. GCNs can, in a broader sense, also be classified as NRL methods when regarding the hidden layers of nodes as embedding vectors. However, few existing NRL models incorporate link attributes. In Section 5, We compare the performances of LASE with several other NRL approaches.

\begin{table}
\centering
\begin{tabular}{l|l}
\toprule
$G=(V,E)$       & Input graph.     \\
\midrule
$u,v,w,\cdots$  & Nodes in G.   \\
\midrule
$e_{u,v}$       & A link from node $u$ to $v$. \\
\midrule
$(v, e_{u,v})$  & A pair of node and link, or to say a \\
                & {\it neighbor} of node $u$. \\
\midrule
$f(\cdot)$      & The feature of a node, a link or a pair. \\
\midrule
$N(u)$          & The set of neighbor nodes of $u$. \\
\midrule
$\otimes,\odot,\langle\cdot,\cdot\rangle$
                & The operations of tensor product, elem- \\
                & -ent-wise product and inner product. \\
\midrule
$[\cdot,\cdots,\cdot],[\vdots],\oplus$
                & The operation of concatenating input \\
                & vectors. \\
\midrule
$\mathbf{W}, \mathbf{U}, \mathbf{V},\cdots$
                & The parameters in the neural network.\\
\midrule
$h^{(i)}(u)$    & The hidden representation of node $u$ in  \\
                & layer $i$. \\
\bottomrule
\end{tabular}
\caption{Notations in this paper.}
\label{tab1}
\end{table}

\section{Model}

\begin{figure}
  \centering
  \includegraphics[width=0.95\linewidth]{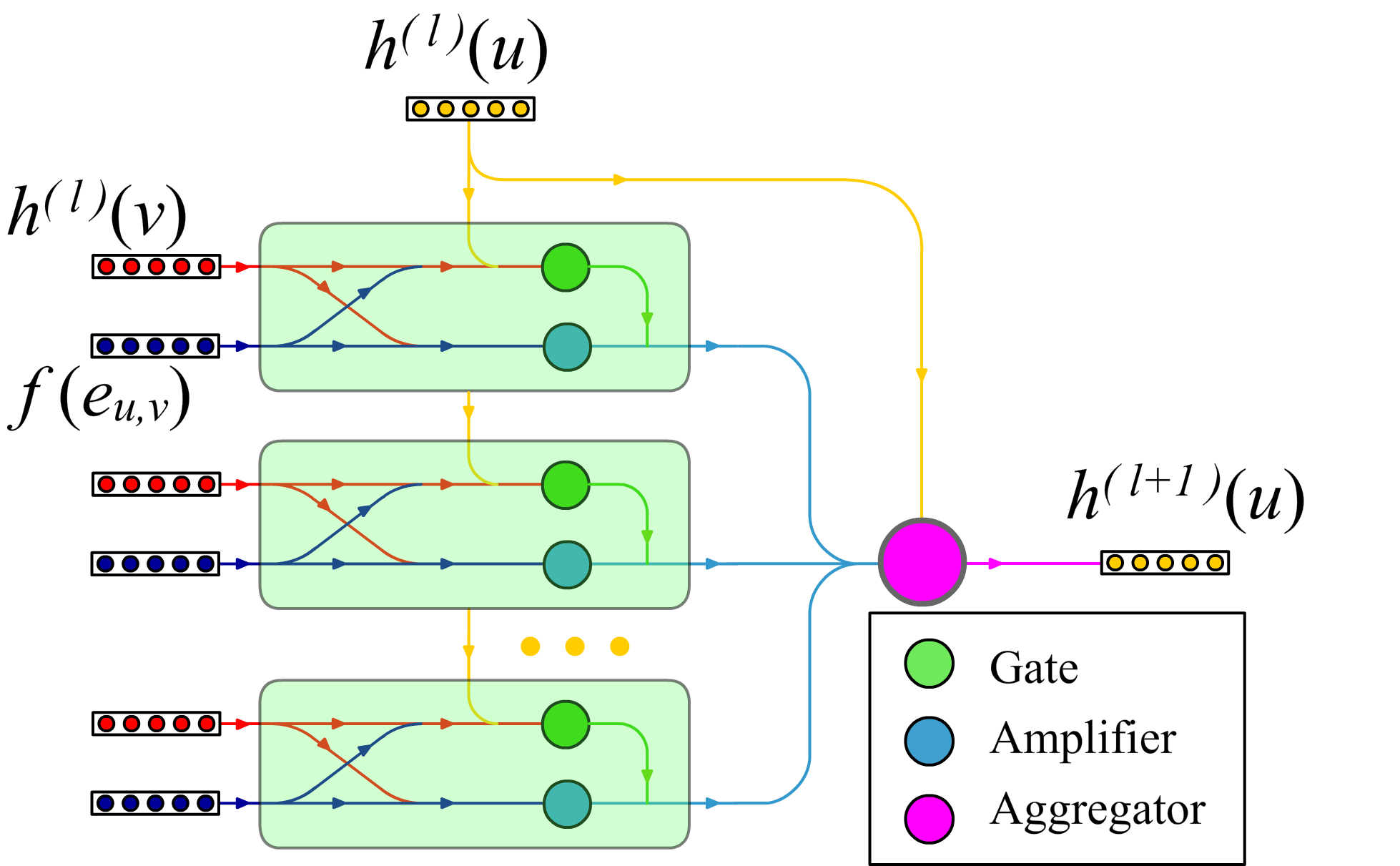}
  \caption{The architecture of LASE.}\label{fig2}
\end{figure}

  In this section, we introduce the architecture of LASE (see Figure~\ref{fig2}) and the motivation behind. We first extend the concept of a {\it neighbor} \footnote{The term {\it neighbor} is used for an ordered pair, containing a neighbor node and the link connecting it to the central node, similarly hereinafter.}, and the features of a neighbor is then defined as the {\it tensor product} of its node and link attributes. As directly using these tensors of features in GCNs would be clumsy, we design new graph kernels under this setup, based on which we further derive possible architectures of LASE. In the end, we modularize LASE and provide intuitive understandings of the modules' functions. Notations in this paper are illustrated in Table~\ref{tab1}. Vectors in this paper are all formed in columns.

\subsection{Neighbor Feature Tensors}

\begin{figure}
  \centering
  \includegraphics[width=0.36\textwidth]{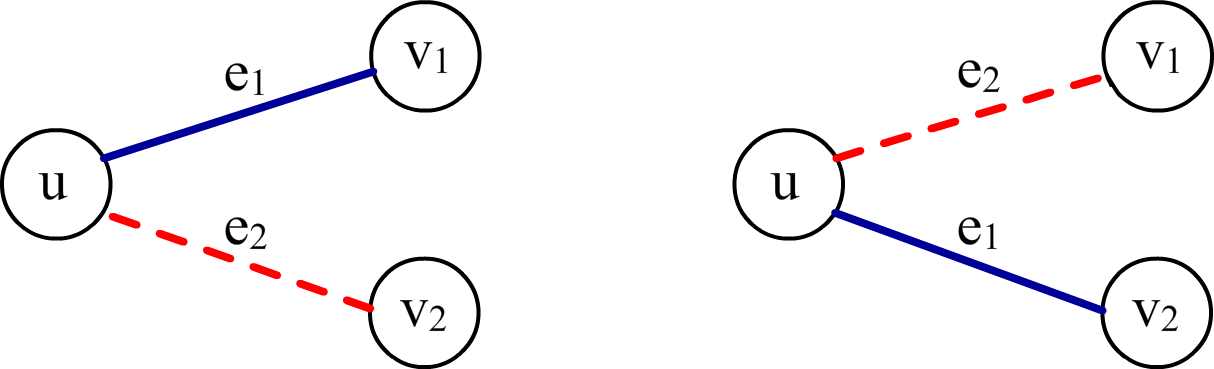}
  \caption{Simply concatenating node and link attributes does not capture the interactions in between. It cannot distinguish between the two situations above.}\label{fig3}
\end{figure}

  One simplest idea to incorporate link attributes in GCN models is to concatenate them to the node's attributes, i.e.
\begin{align*}
    h^{(i+1)}(u) & = \sigma\left(\sum_{v \in N(u)} \mathbf{W}
    \left[
    \begin{array}{c}
      h^{(i)}(v) \\
      f(e_{u,v})
    \end{array}
    \right]\right) \\
     & = \sigma
     \left( \mathbf{W}_1 \sum_{v \in  N(u)} h^{(i)}(v) +
     \mathbf{W}_2 \sum_{v \in N(u)} f(e_{u,v}) \right)
\end{align*}
  where $\mathbf{W}=[\mathbf{W}_1,\mathbf{W}_2]$. However, as attributes of neighbor nodes and neighbor links are individually summed among the neighborhood, such implementations can not at all capture the interactions within a $(node, link)$ neighbor. As the key idea behind LASE is to {\it adequately} incorporate link attributes into node hidden representations, these interactions cannot be ignored. Moreover, this setup also leads to the confusion demonstrated in Figure~\ref{fig3}, indicating that the graph structure is not appropriately captured.

  Instead of simply adding or concatenating node and link attributes, we define their {\it tensor product}\footnote{Briefly speaking, the tensor product of two (column) vector $a$ and $b$ is calculated as $ab^T$ with shape $(d_a \times d_b)$.} as the associated neighbor feature. The calculated tensor contains all bilinear combinations of the two attributes, and serves as a fully conjoined feature. Formally, for the central node $u$ connected to $v$ by a link $e_{u,v}$ , the corresponding neighbor feature is defined as
\begin{displaymath}
  f((v, e_{u, v})) := f(v) \otimes f(e_{u, v}).
\end{displaymath}
   However, directly using the tensor as inputs leads to unacceptably high dimensionalities and heavy redundancies, for the feature matrix ({\it i.e. the 2D-tensor}) is only of rank $1$. Instead, we adapt existing graph kernels for so-defined neighbor features, and derive a concise architecture for LASE following \cite{ker2nn}'s route.

\subsection{Graph Kernels with Link Attributes}

  To adapt existing kernels to our setup of neighbor features, we first define the kernel of two neighbors, $(v, e_{\cdot, v})$ and $(w, e_{\cdot, w})$. The {\it neighbor kernel} is defined as the inner product of the neighbor tensors, i.e.
\begin{align*}
  \mathcal{K}((v, e_{\cdot, v}), (w, e_{\cdot, w})):= &
  \langle f((v, e_{\cdot, v})), f((w, e_{\cdot, w})) \rangle \\
  = &\langle f(v), f(w) \rangle \cdot \langle f(e_{\cdot, v}), f(e_{\cdot, w}) \rangle.
\end{align*}
  Based on the neighbor kernel, a kernel of two $l$-hop neighborhoods with central node $u$ and $u'$ can be defined as $\mathcal{K}^{(l)}_{N}(u,u'):=$
\begin{displaymath}
  \left\{
  \begin{array}{lr}
    \langle f(u), f(u') \rangle & l=0 \\
    \langle f(u), f(u') \rangle \cdot \lambda \cdot & \\
    \qquad \sum\limits_{v\in N(u)}\sum\limits_{v' \in N(u')}\mathcal{K}^{(l-1)}_{N}(v,v') \cdot & \\
    \qquad \qquad \langle f(e_{u,v}), f(e_{u',v'}) \rangle & l>0
  \end{array}
  \right.
\end{displaymath}
  by regarding the lower-hop kernel, $\mathcal{K}^{(l-1)}_{N}(v,v')$, as the inner product of the $(l-1)$-th hidden representations of $v$ and $v'$. $\lambda \in (0,1)$ is a decay factor. Furthermore, by recursively applying the neighborhood kernel, we can derive the $l$-hop Random Walk kernel for {\it graphs with link attributes} as
\begin{align*}
  \mathcal{K}^{l}_{W}(G,G') & =
    \lambda^{l-1} \sum_{\mathbf{u} \in P_l(G)} \sum_{\mathbf{u}' \in P_l(G')} \left( \prod_{i=0}^{l-1} \langle f(u_i), f(u'_i) \rangle \right.\\
  & \quad \left. \times \prod_{i=0}^{l-2} \langle f(e_{u_i,u_{i+1}}), f(e_{u'_i,u'_{i+1}}) \rangle \right),
\end{align*}
  where $P_l(G)$ denotes the set of all walk sequences of length $l$ in graph $G$, and $u_i$ denotes the $i$-th node in sequence $\mathbf{u}$. \footnote{A Weisfeiler-Lehman kernel can also be defined by adopting the {\it graph relabeling} process, which is detailedly introduced in \cite{ker2nn}. We skip this part due to the limitation of space.}

\subsection{From Kernels to Neural Architectures}

  Following the route introduced in \cite{ker2nn} with above Random Walk kernel, a corresponding architecture {\it LASE-RW} can be immediately derived as
\begin{align*}
  \lambda^{(l)}_{u,v} &=
    \sigma( \mathbf{V}^{(l)}[h^{(l)}(u),f(e_{u,v}),h^{(l)}(v)]
      +\mathbf{b^{(l)}} ) \\
  h^{(0)}(u) &= \mathbf{W}^{(0)} f(u) \\
  h^{(l)}(u) &=
    \sum_{v \in N(u)} \lambda_{u,v} h^{(l-1)}(u)
      \odot\mathbf{U}^{(l)}f(e_{u,v})
        \odot\mathbf{W}^{(l)}f(v).
\end{align*}
 This architecture further enjoys a similar property to the architecture described in \cite{ker2nn}. We first construct $L_{l,k}$ $=$ $(V_L,E_L)$ using the $k$-th row vectors in the parameter matrices\footnote{Denoted as $\{\mathbf{W}_k^{(0)},$ $\cdots,$ $\mathbf{W}_k^{(l)}\}$ and $\{\mathbf{U}_k^{(1)},$ $\cdots,$ $\mathbf{U}_k^{(l)}\}$.} as node and link attributes. The node set $V_L$ $=$ $\{v_0,$ $\cdots,$ $v_l\}$ with $f(v_i)$ $=$ $\mathbf{W}_k^{(i)}$, and the link set $E_L$ $=$ $\{e_{v_i,v_{i+1}}\}$ with $f(e_{v_i,v_{i+1}})$ $=$ $\mathbf{U}_k^{(i+1)}$. Then, we have the following theorem:

  \paragraph{Theorem 1.} {\it For any $l$ $\ge$ $1$, the sum of $h^{(l)}(v)[k]$ (the $k$-th coordinate of $h^{(l)}(v)$) satisfies
\begin{displaymath}
  \sum_{v \in V_G}h^{(l)}(v)[k]=\mathcal{K}_W^{l}(G,L_{l,k}),
\end{displaymath}
  and thus $\sum_{v \in V_G}h^{(l)}(v)[k]$ lies in the RKHS of kernel $\mathcal{K}_W^{l}$.}

  Similarly, the architecture {\it LASE-WL} derived from Weisfeiler-Lehman Kernel should be adapted as
\begin{align*}
  \lambda^{(l)}_{u,v} &=
    \sigma(\mathbf{V}^{(l)}[h^{(l)}(u),f(e_{u,v}),h^{(l)}(v)]
      +\mathbf{b}^{(l)}) \\
  h^{(0)}(u) &= \mathbf{W}^{(0)} r^{(d-1)}(u) \\
  \nonumber h^{(l)}(u) &=
    \sum_{v \in N(u)}\lambda_{u,v}h^{(l-1)}(v) \\
      & \quad \odot \mathbf{U}^{(l)} f(e_{u,v})
        \odot \mathbf{W}^{(l)} r^{(d-1)}(u) \\
  \nonumber r^{(d)}(u) &=
    \sigma(\mathbf{P}_1 r^{(d-1)}(u) + \mathbf{P}_2 \sum_{v \in N(u)} \sigma(\mathbf{Q} r^{(d-1)}(v)) )
\end{align*}

  The Weisfeiler-Lehman architecture is originally designed to convolute nodes through both depth ($r$) and breadth ($h$), however, the calculation of LASE-WL would be too complex. We unite the depth and breadth convolution to reduce model size, and by referring to the neighborhood aggregation concept in GraphSAGE \cite{sage}, proposed LASE-SAGE:
\begin{align*}
  \lambda^{(l)}_{u,v} &= \sigma(\mathbf{V}^{(l)}[h^{(l)}(u), f(e_{u,v}), h^{(l)}(v)] + \mathbf{b^{(l)}}) \\
  h^{(0)}(u) &= f(u) \\
  h^{(l)}(u) &= \sigma\left(\mathbf{W}^{(l)}_1 h^{(l-1)}(u) \circ \mathbf{W}^{(l)}_2 \cdot \right. \\
  & \qquad \left.\sum_{v \in N(u)} \lambda_{u,v} h^{(l-1)}(v) \odot \mathbf{U}^{(l)} f(e_{u,v})\right),
\end{align*}
where $\circ$ is a combination operation such as $+$, $\odot$ or $\oplus$.

\subsection{Discussion}

  Although \cite{ker2nn} is originally introduced for aggregating features for entire graphs, its output graph features are an activated sum of all nodes features. We reckon these node features be as well informative in node-wise prediction tasks. We also provide an intuitive understanding of LASE. The calculations in LASE can be divided into three common modules, namely a {\it gate}, an {\it amplifier} and an {\it aggregator}, as is shown in Figure~\ref{fig2}. Intuitively, the gate ($\lambda_{u,v}$) evaluates $v$'s {\it influence} in $u$'s neighborhood. The amplifier ($h(v) \odot \mathbf{U} f(e_{u,v})$) element-wisely {\it amplifies} the node attributes using link information \footnote{A slight elevation in performance can be observed when applying a sigmoid activation on $\mathbf{U} f(e_{u,v})$, which makes the vector functions more analogously to an amplifier.}. The aggregator sums up neighbor embeddings and combines them with the central node embedding using various strategies. Aggregators proposed in \cite{sage} may also be used in LASE.

\section{Sampling Estimation}

  Similar to GCN \shortcite{kipf-gcn}, scalability is an obvious challenge for LASE: calculating the convolutions demands a recursively expanded neighborhood. For nodes with high degrees, it will quickly cover a large portion of the graph. To control batch scales, we leverage the Monte Carlo method to estimate the summed neighborhood information by sampling a fixed number of neighbors. Despite different architectures, the output hidden embeddings of LASE can all be formulated as
\begin{align*}
  h^{(l)}(u) &= \sigma_*\left(
    \sum_{v \in N(u)} \lambda^{(l)}_{u,v} g^{(l)}(v|u)
  \right) \\
  &= \sigma_*\left(
    \mathbb{E}_{p^{(l)}(\cdot|u)}
    \left[\frac{\lambda^{(l)}_{u,v} g^{(l)}(v|u)}{p^{(l)}(v|u)}\right]
  \right)
\end{align*}
where $p^{(l)}(\cdot|u)$ denotes the sampling probabilities in $N(u)$. We then approximate $h^{(l)}(u)$ through estimating the expectation. As the sampling process is always unbiased, we look for the optimal probabilities that minimize the estimation variance.

Although there are existing sampling strategies proposed for GCNs (\cite{fastgcn} \cite{ada-sample}), these methods cannot be directly transferred to LASE because of the absence of explicit, constant link weights. Besides, the optimal distribution varies through the training process. However, a similar idea of importance sampling, coined {\it gate sampling}, can be used in LASE by regarding the decay factor $\lambda$s as the sampling weights, that is,
\begin{equation*}
  p^{(l)}_{gate}(v|u) = \frac{\lambda^{(l)}_{u,v}}{\sum_{w \in N(u)} \lambda^{(l)}_{u,w}}.
\end{equation*}
While sampling with gates may reduce the estimation variance, it is not an optimal solution because typically the norms of $g^{(l)}(v|u)$s are different. According to the derivations of importance sampling in \cite{mcs}, we derive {\it min\_var sampling}, the optimal sampling probabilities as
\begin{equation*}
  p^{(l)}_*(v|u) =
  \frac{\lambda^{(l)}_{u,v} \Vert g^{(l)}(v|u) \Vert}
  {\sum_{w \in N(u)}
      \lambda^{(l)}_{u,w} \Vert g^{(l)}(w|u) \Vert}.
\end{equation*}

Evaluating the sampling probabilities batch-wisely can be rather inefficient. Under the hypothesis that the network parameters do not dramatically vary from batch to batch, we make a tradeoff between variance and efficiency by controlling the interval of calculating the optimal distribution. That is, the sampling probabilities for all training nodes are calculated every $k$ batches. Although the calculation may be time-consuming, the batch-averaged time cost will be reduced to $1/k$.

\section{Experiments}

\subsection{Experiment Setups}

\paragraph{Datasets.} We validate our method on the four datasets introduced below, including a {\it graph with link attributes} ({\tt reddit}), a {\it graph with heterogenous links} ({\tt dblp}), and two {\it temporal networks} ({\tt email}, {\tt fmobile}). The statistics of datasets are shown in Table~\ref{tab2}.

\begin{itemize}
  \item {\tt Reddit} is a Reddit post network with each node representing a post, and each link indicating that the connected two posts are commonly commented by at least three users. We adopt the same setup as \cite{sage} for the node attributes, and use the user-averaged distributions of comments in different communities as link attributes.
  \item {\tt Dblp} is a co-author network constructed with papers from 2013 to 2017 in eight artificial intelligence conferences. We use the {\it tf-idf} vectors of paper titles as node attributes. The links are categorized under author perspective, i.e. the one-hot embeddings of the common authors are used as link attributes. The node and link attributes are reduced to 200 dimensions using PCA.
  \item {\tt Email} and {\tt fmobile} are two temporal networks constructed with user contacts in email and mobile-phone services. The contacts of exact times are discretized into time-slices and used as link attributes. As there is no available node features in the datasets, we use node embeddings with {\it dim=128} obtained from transductive-LINE \shortcite{line} as the pre-trained node features in all convolution-based models.
\end{itemize}

\begin{table}
  \centering
  \begin{tabular}{lrrrr}
    \toprule
    Datasets & $n_{nodes}$ & $n_{links}$ & $\bar{d}$ & $n_{labels}$\\
    \midrule
    {\tt reddit} & 61,836 & 1,222,411 & 19.77 & 8 \\
    {\tt dblp} & 14,389 & 111,858 & 7.77 & 8 \\
    {\tt email} & 986 & 16,064 & 16.29 & 42 \\
    {\tt fmobile} & 21,102 & 55,009 & 2.61 & 33 \\
    \bottomrule
  \end{tabular}
  \caption{Statistics of datasets used in this paper.}
  \label{tab2}
\end{table}

\paragraph{Baselines.} We compare the performance of LASE with baselines including raw features, LINE \shortcite{line}, DeepWalk \shortcite{dw}, GCN \shortcite{kipf-gcn} and GraphSAGE \shortcite{sage}. For LINE and DeepWalk, we adopt an online-style training strategy for the test / validation set introduced in \cite{sage} \footnote{As there is no implementation of online-LINE and \cite{netmf} proves that LINE is theoretically equivalent with DeepWalk with {\it walk\_length=1}, we use the implementation of online-DeepWalk in \cite{sage} instead. {\it N\_walks} is respectively added to compensate the reduction in node contexts.}, and a one-layer softmax-activated neural classifier is trained for all models. To demonstrate the ability of LASE in leveraging link attributes through the {\it amplifiers}, we also test the performance of a LASE variant, {\it LASE-concat}, implemented by na\"ively concatenating link attributes to node attributes.

\subsection{Node-wise Classification}

We implement node-wise classification respectively over the four datasets mentioned above by predicting the community ({\tt reddit}, {\tt email} and {\tt fmobile}) or the conference ({\tt dblp}) that a node belongs to. In all datasets, $65\%$ nodes are used as the training set, $15\%$ as the validation set and the rest as the test set. The nodes in the training set with no neighbors are abandoned. The {\it micro-averaged f1 scores} on the test set are shown in Table~\ref{tab3}.

As one of the most distinguished strengths of GCNs is to aggregate neighborhood features, convolutional-based models including GCN, GraphSAGE and LASE show significant advantages to proximity-based models on datasets with node attributes. Through leveraging link attributes, LASE outperforms other GCNs. Moreover, with LASE-RW and LASE-SAGE outperforming the na\"ive implementation LASE-concat, the effect of the {\it amplifier} module can be corroborated. Although there is no original features in two temporal networks, LASE still outperforms pre-trained features by exploring edge attributes, while GCN and GraphSAGE do not capture these additional information and struggles in over-fitting the proximity-based features.

\begin{table*}
  \centering
  \resizebox{440pt}{!}{
  \begin{tabular*}{460pt}{@{\extracolsep{\fill}}p{100pt}|rrrr}
    \toprule
    Model & {\tt reddit} & {\tt dblp} & {\tt email} & {\tt fmobile} \\
    \midrule
    LINE (online)     & 0.1802 & 0.2989 & 0.3604* & 0.3047* \\
    DeepWalk (online) & 0.1714 & 0.3306 & 0.3249* & 0.4071* \\
    GCN               & 0.8172  & 0.5033 & 0.6396 & 0.3908 \\
    GraphSAGE         & 0.8468 & 0.5798 & 0.6548 & 0.5334 \\
    LASE-concat       & 0.8438 & 0.5805 & 0.7005 & 0.5380 \\
    LASE-RW           & 0.8460 & 0.5433 & 0.7208 & 0.5441 \\
    LASE-SAGE         & {\bf 0.8633} & {\bf 0.5881} & {\bf 0.7310} & {\bf 0.5649} \\
    \midrule
    Raw Features      & 0.7923 & 0.4532 & - & - \\
    LINE (transd.)    & - & - & 0.6904 & 0.4749 \\
    \bottomrule
  \end{tabular*}}
  \caption{Performances of node-wise prediction tasks of LASE, its variants and baselines (Micro-f1s). *: Comparisons between these results and those of convolutional models would be considered unfair as the latter uses transductively learned features as inputs.} \label{tab3}
\end{table*}

\begin{figure*}
  \centering
  \includegraphics[width=0.81\textwidth]{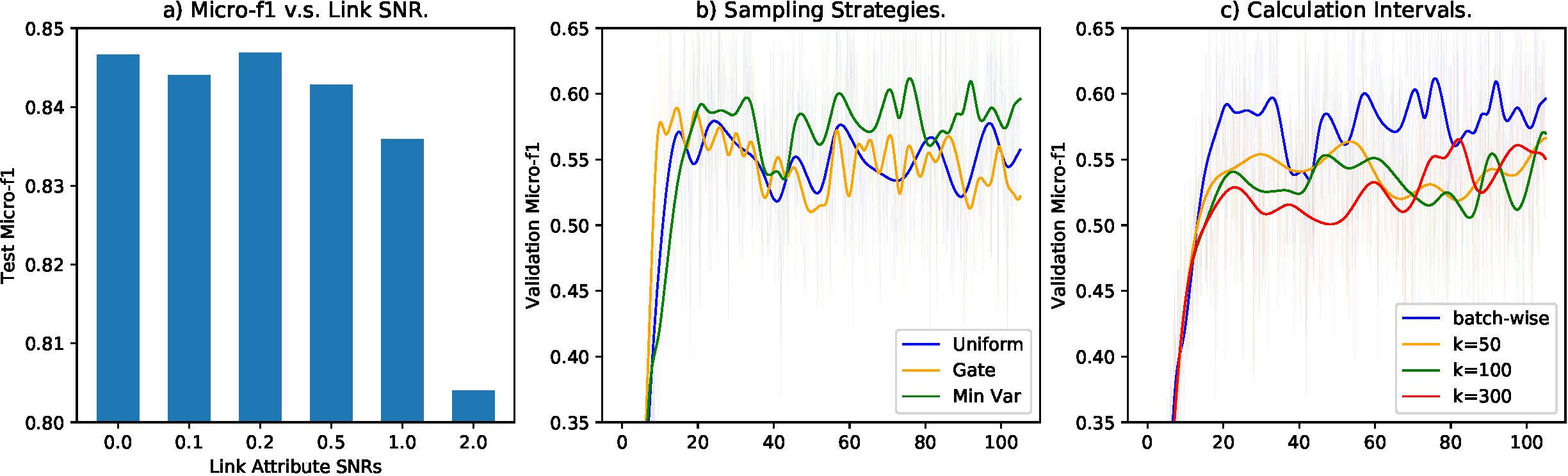}
  \caption{Analyses of LASE: {\bf a)} the prediction accuracies with contaminated link attributes. With higher SNR, the test accuracy of LASE on {\tt reddit} significantly drops; {\bf b)} the validation accuracies on {\tt email} with different sampling strategies; {\bf c)} the validation accuracies on {\tt email} with different intervals of calculating the optimal sampling probabilities for {\it min\_var}. }\label{fig4}
\end{figure*}

  Figure~\ref{fig4} {\bf a)} demonstrate the accuracies of LASE-SAGE using contaminated link attributes of different signal-to-noise ratios (SNRs). That is, we add normal-distributed noises of different standard deviations to the original link attributes according to given SNRs, and separately train LASE-SAGE models under identical model settings. The SNR is defined as
\begin{displaymath}
    SNR = A(link\_attrs) / A(noises)
\end{displaymath}
  where $A(\cdot)$ denotes the standard deviation of the inputs. As SNR increases, a significant trend of decrease in accuracy can be observed. This corroborates the informativeness of link attributes in the LASE architecture.

\subsection{Comparison of Sampling Strategies}

  We look into the training processes of different neighborhood sampling strategies introduced in Section 4, namely {\it uniform}, {\it gate} and minimal variance ({\it min\_var}) sampling. We separately train models with corresponding sampling strategy on {\tt email}, and present the variations of accuracies on the validation set against training epochs in Figure~\ref{fig4} {\bf b)}. While the convergence speeds appear analogous, {\it min\_var} sampling consistently attains better convergence performance compared with uniform and gate sampling. The reason that {\it gate} sampling does not show a significant advantage over {\it uniform} sampling may be that the norms of transformed neighbor features ($g^{(l)}(v|u)$) varies greatly in the neighborhood.

  Figure~\ref{fig4} {\bf c)} shows the tradeoff between performance and efficiency made through different calculation intervals of the sampling distribution (under {\it min\_var} setup). As the interval $k$ increases, the performance slightly drops. Calculating the probabilities batch-wise attains a significant elevation in the performance, while the computation cost can be unacceptably high on larger datasets. Additionally, when $k$ becomes large enough, increasing $k$ does not significantly influence the training performance.

\section{Conclusions and Future Work}

In this paper, we propose LASE as an extension of graph convolutional networks, which leverages more information from graph links than existing GCNs by incorporating the link attributes. The contribution of LASE lies in three folds: i) LASE provides a ubiquitous solution to a wider class of graph data by incorporating link attributes; ii) LASE outperforms strong baselines and na\"ive concatenating implementations by adequately leveraging the information in the link attributes; iii) LASE adopt a more explainable approach in determining the neural architecture and thus enjoys better explainability.

For future work, we are looking for better sampling solutions for LASE, as although stressed with calculation intervals, current sampling setup seems to be rather clumsy when the graph becomes massively large. We are also looking for other possible approaches, hopefully with better performance, to incorporating link attributes. Besides, as LASE is an universal solution to all graph-structured data, an intriguing direction may be designing domain- or task-specific architectures based on LASE to attain better performances, such as more elegant adaptations to dynamic networks.

\bibliographystyle{named}
\bibliography{ijcai19}

\begin{thebibliography}{}

\bibitem[\protect\citeauthoryear{Bruna \bgroup \em et al.\egroup
  }{2014}]{bruna-gcn}
Joan Bruna, Wojciech Zaremba, Arthur Szlam, and Yann Lecun.
\newblock Spectral networks and locally connected networks on graphs.
\newblock In {\em Proceedings of the 2nd International Conference on Learning
  Representations (ICLR'14)}, 2014.

\bibitem[\protect\citeauthoryear{Du \bgroup \em et al.\egroup }{2018}]{dne}
Lun Du, Yun Wang, Guojie Song, Zhicong Lu, and Junshan Wang.
\newblock Dynamic network embedding : An extended approach for skip-gram based
  network embedding.
\newblock In {\em Proceedings of the 27th International Joint Conference on
  Artificial Intelligence (IJCAI'18)}, 2018.

\bibitem[\protect\citeauthoryear{G\"{a}rtner \bgroup \em et al.\egroup
  }{2003}]{rw-kernel}
Thomas G\"{a}rtner, Peter Flach, and Stefan Wrobel.
\newblock On graph kernels: Hardness results and efficient alternatives.
\newblock In {\em Proceedings of the Annual Conference on Computational
  Learning Theory (CoLT'03)}, 2003.

\bibitem[\protect\citeauthoryear{Grover and Leskovec}{2016}]{n2v}
Aditya Grover and Jure Leskovec.
\newblock node2vec: Scalable feature learning for networks.
\newblock In {\em Proceedings of the 22nd ACM SIGKDD International Conference
  on Knowledge Discovery and Data Mining (KDD'16)}, pages 855--864, 2016.

\bibitem[\protect\citeauthoryear{Hamilton \bgroup \em et al.\egroup
  }{2017}]{sage}
William~L. Hamilton, Rex Ying, and Jure Leskovec.
\newblock Inductive representation learning on large graphs.
\newblock In {\em Proceedings of the 31st Conference on Neural Information
  Processing Systems (NeurIPS'17)}, 2017.

\bibitem[\protect\citeauthoryear{Henaff \bgroup \em et al.\egroup
  }{2015}]{henaff-gcn}
Mikael Henaff, Joan Bruna, and Yann Lecun.
\newblock Deep convolutional networks on graph-structured data.
\newblock {\em Computer Science}, 2015.

\bibitem[\protect\citeauthoryear{Huang \bgroup \em et al.\egroup
  }{2018}]{ada-sample}
Wen{-}bing Huang, Tong Zhang, Yu~Rong, and Junzhou Huang.
\newblock Adaptive sampling towards fast graph representation learning.
\newblock {\em CoRR}, abs/1809.05343, 2018.

\bibitem[\protect\citeauthoryear{Jie \bgroup \em et al.\egroup
  }{2018}]{fastgcn}
Chen Jie, Tengfei Ma, and Xiao Cao.
\newblock Fastgcn: Fast learning with graph convolutional networks via
  importance sampling.
\newblock In {\em Proceedings of the 6th International Conference on Learning
  Representations (ICLR'18)}, 2018.

\bibitem[\protect\citeauthoryear{Kipf and Welling}{2016}]{kipf-gcn}
Thomas~N. Kipf and Max Welling.
\newblock Semi-supervised classification with graph convolutional networks.
\newblock In {\em Proceedings of the 4th International Conference on Learning
  Representations (ICLR'16)}, 2016.

\bibitem[\protect\citeauthoryear{Li \bgroup \em et al.\egroup }{2018}]{sepne}
Ziyao Li, Liang Zhang, and Guojie Song.
\newblock Sepne: Bringing separability to network embedding.
\newblock {\em CoRR}, abs/1811.05614, 2018.

\bibitem[\protect\citeauthoryear{Niepert \bgroup \em et al.\egroup
  }{2016}]{ger-gcn}
Mathias Niepert, Mohamed Ahmed, and Konstantin Kutzkov.
\newblock Learning convolutional neural networks for graphs.
\newblock In {\em Proceedings of the 33rd International Conference on Machine
  Learning (ICML'16)}, 2016.

\bibitem[\protect\citeauthoryear{Owen}{2013}]{mcs}
Art~B. Owen.
\newblock {\em Monte Carlo theory, methods and examples}.
\newblock 2013.

\bibitem[\protect\citeauthoryear{Perozzi \bgroup \em et al.\egroup }{2014}]{dw}
Bryan Perozzi, Rami Al-Rfou, and Steven Skiena.
\newblock Deepwalk: online learning of social representations.
\newblock In {\em Proceedings of the 20th ACM SIGKDD International Conference
  on Knowledge Discovery and Data Mining (KDD'14)}, 2014.

\bibitem[\protect\citeauthoryear{Qiu \bgroup \em et al.\egroup }{2018}]{netmf}
Jiezhong Qiu, Yuxiao Dong, Hao Ma, Jian Li, Kuansan Wang, and Jie Tang.
\newblock Network embedding as matrix factorization: Unifying deepwalk, line,
  pte, and node2vec.
\newblock In {\em Proceedings of the 11th ACM International Conference on Web
  Search and Data Mining (WSDM'18)}, 2018.

\bibitem[\protect\citeauthoryear{Ribeiro \bgroup \em et al.\egroup
  }{2017}]{struct2vec}
Leonardo~F.R. Ribeiro, Pedro~H.P. Saverese, and Daniel~R. Figueiredo.
\newblock Struc2vec: Learning node representations from structural identity.
\newblock In {\em Proceedings of the 23rd ACM SIGKDD International Conference
  on Knowledge Discovery and Data Mining (KDD'17)}, 2017.

\bibitem[\protect\citeauthoryear{Sch\"{o}lkopf and Smola}{2002}]{kernel}
B.~Sch\"{o}lkopf and Alexander~Johannes Smola.
\newblock {\em Learning With Kernels}.
\newblock The MIT Press, 2002.

\bibitem[\protect\citeauthoryear{Shervashidze \bgroup \em et al.\egroup
  }{2011}]{wl-kernel}
Nino Shervashidze, Pascal Schweitzer, Erik Jan, Van Leeuwen, Kurt Mehlhorn, and
  Karsten~M. Borgwardt.
\newblock Weisfeiler-lehman graph kernels.
\newblock {\em Journal of Machine Learning Research}, 12(3):2539--2561, 2011.

\bibitem[\protect\citeauthoryear{Tang \bgroup \em et al.\egroup }{2015}]{line}
Jian Tang, Meng Qu, Mingzhe Wang, Ming Zhang, Jun Yan, and Qiaozhu Mei.
\newblock Line:large-scale information network embedding.
\newblock In {\em Proceedings of the 24th International Conference on World
  Wide Web (WWW'15)}, 2015.

\bibitem[\protect\citeauthoryear{Tao \bgroup \em et al.\egroup }{2017}]{ker2nn}
Lei Tao, Wengong Jin, Regina Barzilay, and Tommi Jaakkola.
\newblock Deriving neural architectures from sequence and graph kernels.
\newblock In {\em Proceedings of the 34 th International Conference on Machine
  Learning (ICML'17)}, 2017.

\bibitem[\protect\citeauthoryear{Vishwanathan \bgroup \em et al.\egroup
  }{2008}]{g-kernels}
S.~V.~N. Vishwanathan, Karsten~M. Borgwardt, Imre~Risi Kondor, and Nicol~N.
  Schraudolph.
\newblock Graph kernels.
\newblock {\em Journal of Machine Learning Research}, 11(2):1201--1242, 2008.

\bibitem[\protect\citeauthoryear{Wang \bgroup \em et al.\egroup }{2016}]{sdne}
Daixin Wang, Peng Cui, and Wenwu Zhu.
\newblock Structural deep network embedding.
\newblock In {\em Proceedings of the 22nd ACM SIGKDD International Conference
  on Knowledge Discovery and Data Mining (KDD'16)}, 2016.

\end{thebibliography}

\end{document}